\documentclass[iicol]{sn-jnl}

\usepackage{graphicx}%
\usepackage[export]{adjustbox}
\usepackage{multirow}%
\usepackage{amsmath,amssymb,amsfonts}%
\usepackage{amsthm}%
\usepackage{multirow}
\usepackage{enumitem}
\usepackage{balance}
\usepackage{subcaption}
\usepackage[linesnumbered,ruled,vlined]{algorithm2e}
\usepackage{array}
\usepackage{lipsum}
\usepackage{xcolor}
\usepackage{soul}

\robustify\hl

\newcommand{\PreserveBackslash}[1]{\let\temp=\\#1\let\\=\temp}
\newcolumntype{C}[1]{>{\PreserveBackslash\centering}m{#1}}

\begin{document}

\title[Article Title]{Real-time Recognition of Human Interactions from a Single RGB-D Camera for Socially-Aware Robot Navigation}

\author[1]{\fnm{Thanh Long} \sur{Nguyen}}

\author[1]{\fnm{Duc Phu} \sur{Nguyen}}

\author[1]{\fnm{Thanh Thao} \sur{Ton Nu}}

\author[1]{\fnm{Quan} \sur{Le}}

\author[2]{\fnm{Thuan Hoang} \sur{Tran}}

\author*[1]{\fnm{Manh Duong} \sur{Phung}}\email{duong.phung@fulbright.edu.vn}

\affil[1]{\orgdiv{Undergraduate Faculty}, \orgname{Fulbright University Vietnam}, 
\orgaddress{\street{105 Ton Dat Tien}, \city{Ho Chi Minh City}, \postcode{700000}, \country{Vietnam}}}

\affil[2]{\orgdiv{Faculty of Information Technology}, \orgname{Duy Tan University}, 
\orgaddress{\street{3 Quang Trung}, \city{Da Nang}, \postcode{550000}, \country{Vietnam}}}


\abstract{Recognizing human interactions is essential for social robots as it enables them to navigate safely and naturally in shared environments. Conventional robotic systems however often focus on obstacle avoidance, neglecting social cues necessary for seamless human-robot interaction. To address this gap, we propose a framework to recognize human group interactions for socially aware navigation. Our method utilizes color and depth frames from a monocular RGB-D camera to estimate 3D human keypoints and positions. Principal component analysis (PCA) is then used to determine dominant interaction directions. The shoelace formula is finally applied to compute interest points and engagement areas. Extensive experiments have been conducted to evaluate the validity of the proposed method. The results show that our method is capable of recognizing group interactions across different scenarios with varying numbers of individuals. It also achieves high-speed performance, processing each frame in approximately 4 ms on a single-board computer used in robotic systems. The method is implemented as a ROS 2 package making it simple to integrate into existing navigation systems. Source code is available at \url{https://github.com/thanhlong103/social-interaction-detector}.
}


\maketitle

\section{Introduction}

Social robots play a key role in many applications such as elderly care, home assistant, customer service, and education where they assist, interact, and communicate with humans in a socially intelligent manner. These robots must ensure not only physical safety but also psychological comfort for humans by following social norms. For instance, a robot should avoid disrupting a group conversation when navigating a crowded space as this could be seen as impolite or intrusive. To accomplish this, the robot must not only detect humans but also recognize and interpret their interactions such as conversations, discussions, gatherings, and collaborative activities to adapt its movements accordingly.

According to \cite{barua2024enabling, 8036225}, human group interactions are structured into three distinct spaces: (i) o-space, the central region where active participants focus their attention, (ii) p-space, the surrounding area occupied by engaged individuals, and (iii) r-space, the outer region where bystanders or non-participants are positioned. To enable socially aware navigation, recognition algorithms must estimate these spatial regions. Approaches to this problem fall into two categories: model-based \cite{setti2015f,kong2015close,truong2017toward,BIBI2018282} and non-model-based techniques \cite{shu2019hierarchical,9412538, 8892463, nour2014human}. 

The model-based approach extracts the state descriptors for each individual and then infers interactions by analyzing coherence among participants. In \cite{setti2015f}, individuals in the vicinity of the robot are considered for detecting group interactions. A graph-cut method is then used where each individual is represented as a node and edges are defined based on spatial positions and orientation information. An objective function that considers the distances between individuals, visibility constraints, and number of groups is finally defined to determine group interactions. \cite{truong2017toward} extends this approach by incorporating the movement information of each individual to make it more suitable for dynamic environments. In another approach, person-to-person interaction is identified based on features extracted between a focal person and nearby individuals within a predefined distance threshold \cite{BIBI2018282}. It considers action labels, velocities, relative distances, and collective poses to distinguish between interaction types like walk together, stand together, or talking. In \cite{kong2015close}, a patch-aware model is introduced to determine both individual actions and the interaction class by identifying key regions for each person. It treats these regions as hidden variables and uses a discriminative function that incorporates features, individual actions, and interaction labels to evaluate correlation between these elements. While the model-based approach is computationally efficient, it relies on the design of the objective function to identify interactions. Its performance is largely dependent on the quality of this function.

In contrast, the non-model-based approach utilizes deep learning techniques, particularly long short-term memory (LSTM), convolutional neural networks (CNNs), and recurrent neural networks (RNNs), to recognize interactions through labeled data. In \cite{shu2019hierarchical}, a hierarchical long short-term concurrent memory (H-LSTCM) model that captures inter-related dynamics among multiple individuals is introduced to recognize human interactions. This approach first models individual motion using single-person LSTMs, then captures inter-person interactions through a concurrent LSTM (Co-LSTM). In \cite{9412538}, a two-stream recurrent neural network is proposed to recognize human interactions from skeletal sequences by capturing both explicit and implicit spatial relationships between joints. One stream uses pairwise joint distances to model distance patterns, while the other combines these with spatial features to extract joint correlations. An entropy-based fusion strategy is then used to combine the outputs of both streams to identify interactions. A hybrid recognition method that integrates deep learning with a traditional hidden Markov model (HMM) is presented in \cite{8892463}. It first extracts behavior features using an optimized AlexNet and then processes them through an LSTM network with Softmax classification. The final recognition result is obtained by fusing these outputs with HMM results using particle swarm optimization. Other methods employ co-occurring visual words \cite{nour2014human}, interactive phrases \cite{kong2014interactive}, and feature-based convolutional neural network (FCNN) models \cite{verma2021multiperson} to determine group interactions. Nevertheless, challenges remain in generalizing these learning-based approaches across diverse contexts and dynamic environments, as well as in ensuring real-time processing efficiency \cite{mavrogiannis2023core}.

Apart from data processing approach, the choice of sensors also plays a key role in human interaction recognition and socially-aware navigation. LiDAR and cameras are most commonly used as they provide accurate distance measurements and rich visual data for recognition. In \cite{rios2015, 10896312}, data from these sensors are fused to enhance perception reliability in complex social settings. In \cite{li2022deepfusion}, deep features from LiDAR and cameras are combined to improve 3D object detection accuracy in human environments. \cite{tamas2010lidar} demonstrated that combining LiDAR-based Gaussian mixture models (GMMs) with vision-based AdaBoost classifiers enhances human detection and reduces false positives in crowded spaces. Recently, RGB-D cameras that capture both color and depth information have emerged as a replacement for traditional color cameras in robot navigation. Methods using these cameras such as anatomical-plane-based representation \cite{ALAZRAI20152346}, joint distance maps \cite{li2017joint}, deep CNNs \cite{8409991}, and 3D CNNs \cite{6165309,7410867} have shown improved reliability in human interaction recognition. However, these approaches still face limitations in achieving real-time performance.

From the literature survey, it is evident that significant challenges remain in recognizing human interactions for robot navigation, especially in complex scenarios involving multiple interacting individuals. An effective recognition method should be both accurate and simple to implement, ideally requiring only a single sensor device, while also being computationally efficient enough to operate on robots with limited processing capabilities.   

In this work, we propose a novel method for human interaction recognition using a single RGB-D camera. An efficient pretrained neural network is first employed to detect and track human keypoints from the camera’s color stream. The corresponding depth stream is then used to extract the 3D poses of detected individuals. Rather than relying on head orientation, which can be inconsistent and error-prone, our approach utilizes full-body skeleton keypoints to infer human interactions. Inspired by the concept of F-formation \cite{barua2024enabling, yang2017study}, we introduce a geometric method to dynamically estimate interaction zones including both the central interaction area and its surrounding space.

Our contributions are threefold:

\begin{enumerate}[label=(\roman*)]
\item We propose a hybrid approach for human activity recognition that combines deep learning for keypoint detection with mathematical techniques such as principal component analysis (PCA) for human state estimation and group interaction recognition.
\item The hardware setup is minimal, requiring only a single RGB-D camera. The algorithm is robust, capable of recognizing activities even under partial occlusion and operating in real time on resource-constrained platforms like embedded computers in robotic systems.
\item Our method is implemented as a ROS 2 package, making it easy to integrate into existing navigation frameworks and directly enhance a robot's ability to navigate naturally within human environments. The source code is publicly available to support reproducibility and facilitate future research.
\end{enumerate}

The remainder of this paper is organized as follows. Section~\ref{sec2} provides an overview of the proposed system. Section~\ref{sec3} details the estimation of individual human states. Section~\ref{sec4} describes the method for determining group interactions. Experimental results are presented in Section~\ref{sec5}. The paper ends with conclusions drawn in Section~\ref{sec6}.

\section{System Overview}\label{sec2}
\begin{figure*}[!]
    \centering
    \includegraphics[width=0.9\textwidth,frame]{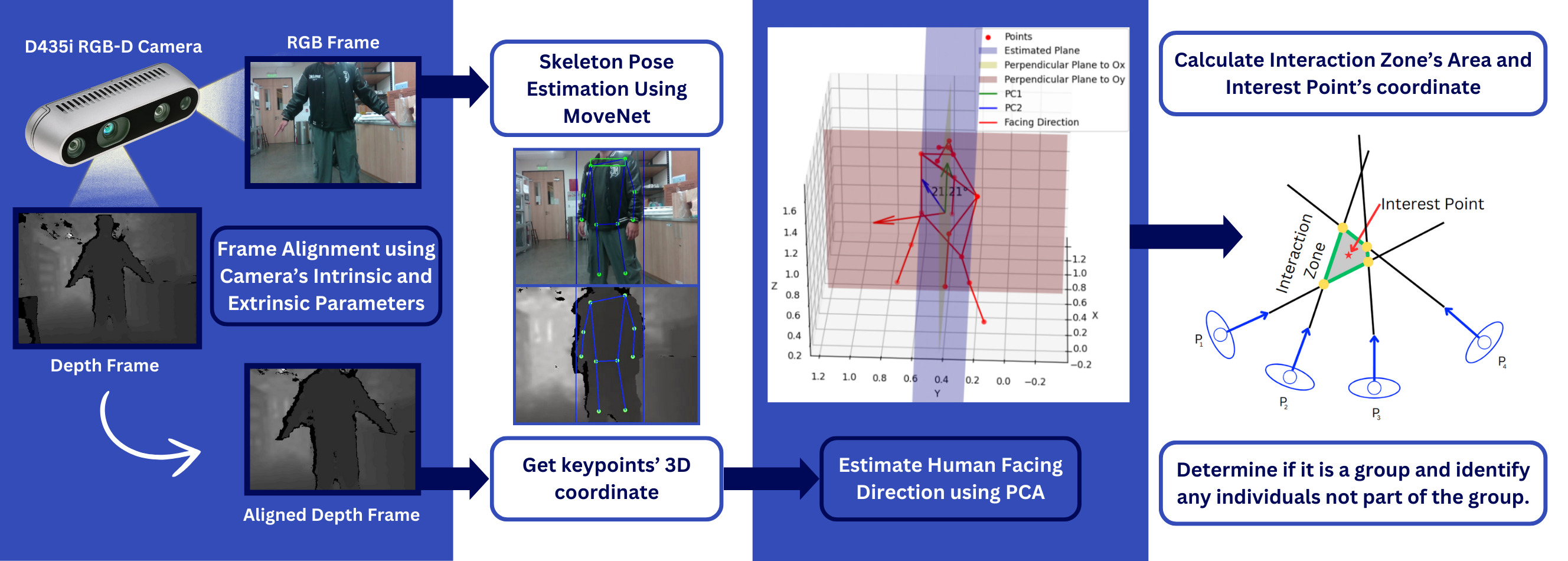}
    \caption{Overview of the proposed human interaction recognition system}
    \label{fig:method}
\end{figure*}

Figure \ref{fig:method} shows a diagram of the proposed human interaction recognition system. The hardware includes a RealSense D435i camera capable of capturing both RGB and depth frames. The RGB frame provides standard color image data, while the depth frame records distance information from the environment. These two frames are then aligned using the camera’s intrinsic and extrinsic parameters. This alignment ensures that the depth data corresponds to the visual information in the RGB image.

Next, the RGB frame is processed by MoveNet \cite{hub2024movenet}, a pre-trained model that identifies keypoints on the human body such as shoulders, hips, elbows, and knees. These 2D keypoints are then mapped to their corresponding depth values using the aligned depth frame, resulting in 3D coordinates for each joint. These coordinates are used to determine the position of each person in the scene.

To estimate the orientation or facing direction of each individual, the system employs principal component analysis (PCA) on the extracted 3D joint positions. PCA identifies the primary direction in which the person’s body is aligned. The final stage focuses on group interaction analysis. Using the facing directions of multiple individuals, the system calculates an interaction zone where their fields of view intersect. Within this zone, the system determines an interest point that represents the focus of group attention. Individuals whose facing directions and spatial positions contribute to this shared focus are classified as part of the group. Details of each stage are presented in the following sections.

\section{Human State Estimation} \label{sec3}
Given input frames from the camera, the system estimates individuals’ state including their position and orientation in 3D space. The estimation begins with RGB frames to identify keypoints in 2D and then uses depth frames to align them to 3D.

\begin{figure*}[!]
    \centering
    \begin{subfigure}[b]{0.35\textwidth}
    \includegraphics[width=\textwidth]{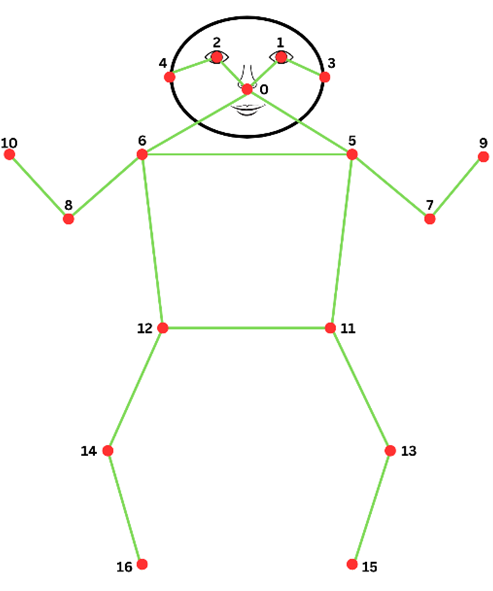}
    \caption{Keypoints definition}
    \label{fig:keypoints}
    \end{subfigure}
    \begin{subfigure}[b]{0.22\textwidth}
    \includegraphics[width=\textwidth]{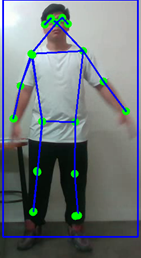}
    \caption{Keypoints detection}
    \label{fig:keypoints_}
    \end{subfigure}
    \caption{Human keypoints for pose estimation}
    \label{fig:keypoints_setup}
\end{figure*}

\subsection{3D Pose Estimation}
For human pose estimation, we utilize MoveNet \cite{hub2024movenet} to detect 17 keypoints $(x_i^{kp},y_i^{kp})$ from the color frame, as shown in Figure \ref{fig:keypoints_setup}. To obtain 3D spatial information, the color frame is aligned with the depth frame so that each keypoint's pixel coordinate in the color frame corresponds to the correct depth measurement in the depth frame.

\subsubsection{Color-Depth Frame Alignment}
Since the color and depth sensors in the camera are physically separated, the depth frame is captured from a slightly different perspective than the color frame. To calibrate it, the depth image is first mapped into the color camera's coordinate space using the camera’s intrinsic parameters. This involves reprojecting each depth pixel $(x^d,y^d,d)$ into 3D camera coordinates $(X^c,Y^c,Z^c )$ using the depth camera’s intrinsic matrix $K_d$:

\begin{equation}
    \begin{bmatrix}
        X^{c} \\
        Y^{c} \\
        Z^{c}
    \end{bmatrix}
    =d.K^{-1}_d
    \begin{bmatrix}
        x^{d} \\
        y^{d} \\
        1
    \end{bmatrix}.
\end{equation}

Since the depth and color sensors have different positions and orientations, we apply an extrinsic transformation $T_{dc}$, which consists of a rotation $R_{dc}$ and a translation $t_{dc}$, to express the 3D point in the color camera's coordinate system:

\begin{equation}
    \begin{bmatrix}
        X^{c}_c \\
        Y^{c}_c \\
        Z^{c}_c
    \end{bmatrix}
    =R_{dc}
    \begin{bmatrix}
        X^{c} \\
        Y^{c} \\
        Z^c
    \end{bmatrix} + t_{dc}.
\end{equation}

The transformed 3D points are then projected back into the color image plane using the color camera's intrinsic matrix $K_c$:

\begin{equation}
    \begin{bmatrix}
        x^{c} \\
        y^{c} \\
        1
    \end{bmatrix}
    =K_c
        \begin{bmatrix}
        X^{c} \\
        Y^{c} \\
        Z^{c}
    \end{bmatrix}.
\end{equation}

After the alignment, the corresponding depth value from the depth frame can be retrieved for each keypoint $(x^{kp},y^{kp})$ detected in the color image. This allows us to compute the full 3D position of each keypoint as:

\begin{equation}
    \begin{bmatrix}
        x_i^{kc} \\
        y_i^{kc} \\
        z_i^{kc}
    \end{bmatrix}
    =d_i^{kp}.K_c^{-1}
        \begin{bmatrix}
        x_i^{kp} \\
        y_i^{kp} \\
        1
    \end{bmatrix}.
\end{equation}

\subsubsection{Transformation to World Coordinates}
Since the robot operates on a 2D ground based on a costmap representing the environment, it is necessary to project the human's position to that costmap. The projection is conducted by first defining the world coordinate system as a top view of the ground, where
    \begin{itemize}
        \item the camera’s depth axis ($z^{kc}$) maps to the world $x-$axis;
        \item the lateral axis ($x^{kc}$) maps to the world $y-$axis.
    \end{itemize}
This transformation can be represented using the following homogeneous transformation matrix:

\begin{equation}
    \mathbf{T}_{cw} =
    \begin{bmatrix}
        0 && 0 && 1 && 0 \\
        1 && 0 && 0 && 0 \\
        0 && 0 && 0 && 1 
    \end{bmatrix}.
\end{equation}

Applying this transformation to each keypoint gives

\begin{equation}
    \mathbf{p}_i^{kw} =  \mathbf{T}_{cw}.\mathbf{p}_i^{kc} = 
    \begin{bmatrix}
        x_i^{kw} \\
        y_i^{kw} \\
        1
    \end{bmatrix}
    =\begin{bmatrix}
        z_i^{kc} \\
        x_i^{kc} \\
        1
    \end{bmatrix}.
\end{equation}

The mean position of all keypoints detected with an acceptable confidence level is used as the final position of a person in the world coordinates, i.e.,

\begin{equation}
    x_i^{p} = \frac{1}{n} \sum_{i=1}^{n} x^{kw}_i, \quad
    y_i^{p} = \frac{1}{n} \sum_{i=1}^{n} y^{kw}_i,
\end{equation}
where $n$ is the number of keypoints that meet the confidence threshold. 

\subsection{Body Direction Estimation}
Body direction is essential for understanding human interaction as it indicates their direction of interest. However, estimating the body direction presents several challenges. A person may be moving in one direction while their head is oriented differently, making head direction alone an unreliable indicator. Furthermore, differentiating whether a person is facing forward or backward based on body posture is non-trivial as symmetric poses may lead to ambiguity. To address these challenges, we propose to estimate the body direction by utilizing all the detected keypoints. The key idea is to analyze the spatial distribution of these keypoints in 3D space to determine the dominant orientation of the human body. We implement this by employing principal component analysis (PCA) to extract a best-fit plane across the detected keypoints. 

\begin{figure*}[!]
    \centering
    \begin{subfigure}[b]{0.45\linewidth}
        \centering
        \includegraphics[width=\linewidth]{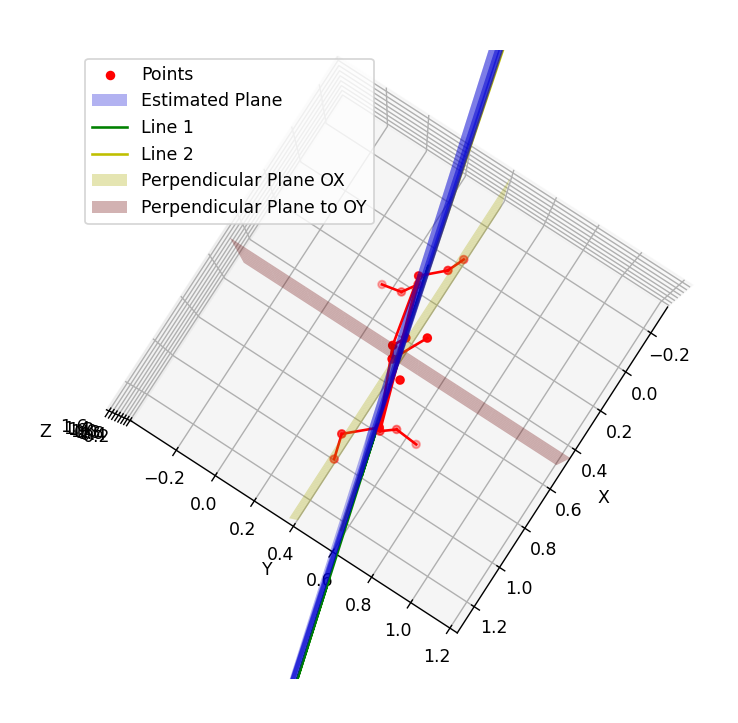}
        \caption{Top View}
        \label{fig:topveiew_plane}
    \end{subfigure}
    \hfill
    \begin{subfigure}[b]{0.45\linewidth}
        \centering
        \includegraphics[width=\linewidth]{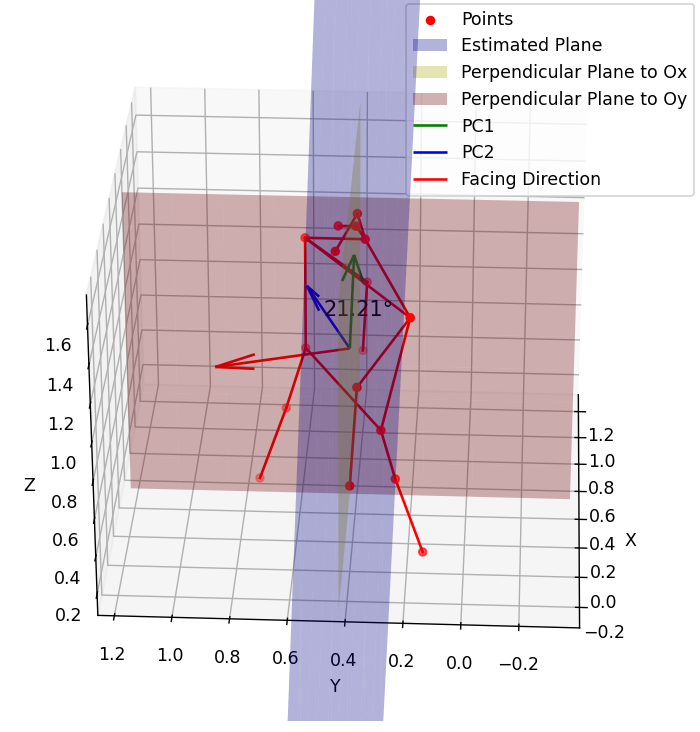}
        \caption{Side View}
        \label{fig:sideview_plane}
    \end{subfigure}
    \caption{Illustration of human pose estimation using principle component analysis (PCA)}
    \label{fig:Plane-visualization}
\end{figure*}

PCA is a statistical technique used to reduce the dimensionality of data while preserving as much variance as possible. It transforms the data into a new coordinate system where the greatest variance lies on the first coordinate (the first principal component). In the context of human pose estimation, PCA is particularly well-suited as it provides a statistical means of determining the primary orientation of a set of points in 3D space. By fitting a plane to the 3D keypoints of the skeleton pose, PCA captures the main orientation of the human body using all available 3D keypoints. The normal vector of this plane serves as a reliable descriptor of the human body's overall orientation. Figure \ref{fig:Plane-visualization} provides a visualization of the fitted plane and its associated normal vector.

Let $K$ denote the set of 3D coordinates of the $n$ keypoints detected for a person:

\begin{equation}
K = \left\{ (x_i^{\text{kpw}},\ y_i^{\text{kpw}},\ z_i^{\text{kpw}})\ \middle|\ i = 1, 2, \ldots, n \right\}.
\end{equation}

Each keypoint contributes to defining the spatial structure of the human body. The collective distribution of these keypoints forms a shape that can be analyzed to determine the body's primary orientation. To eliminate the influence of absolute position, a data matrix $M$ of size $n \times 3$ is constructed where each row contains the $x-$ , $y-$ , and $z-$ coordinates of a keypoint. The matrix is then mean-centered by subtracting the mean position $\overline{M}$:

\begin{equation}
    M_c=M-\overline{M},
\end{equation}
where $\overline{M}=\frac{1}{n} \sum_{i=1}^{n}(x_i^{kpw},y_i^{kpw},z_i^{kpw} )$ . Mean-centering removes positional bias, allowing PCA to extract structural information based purely on body orientation. Next, we apply PCA by computing the covariance matrix $C$ of $M_c$:

\begin{equation}
    C=\frac{1}{n} M_c^T M_c.
\end{equation}

The covariance matrix $C$ captures the variance distribution of the keypoints along different spatial axes. The next step is to compute the eigenvalues $\lambda$ and corresponding eigenvectors $v$ of $C$ that represent the principal axes of variation in the data. This involves solving the characteristic equation: $\det(C - \lambda I) = 0$, where $I$ is the identity matrix of size $3 \times 3$. Solving for $\lambda$ yields three eigenvalues $\lambda_1, \lambda_2, \lambda_3$ such that: $\lambda_1 \geq \lambda_2 \geq \lambda_3 \geq 0$. The corresponding eigenvectors $v_1, v_2, v_3$ define an orthonormal basis aligned with the principal directions of the body’s keypoints. The eigenvector $v_1$ associated with the largest eigenvalue represents the direction of maximum variance, which is usually aligned with the vertical axis of the human body. The second eigenvector $v_2$ corresponds to the primary horizontal direction of the human body. The third eigenvector $v_3$ represents the normal vector $n$ of the best-fitting plane through the keypoints: $n = v_3$. The body orientation is estimated by computing the angle $\theta$ between $n$ and the normal vector $c$ of the camera plane.

\begin{equation}
    \theta_i^p = \arccos\left( \frac{n \cdot c}{\lVert n \rVert \lVert c \rVert} \right)
\end{equation}
This angle provides a general estimate of the body's orientation relative to the camera. However, this alone does not distinguish whether a person is facing forward or backward as the body may appear similar in both cases. To resolve this ambiguity, we utilize the positions of the left and right shoulders by comparing their $x$ coordinates:

\begin{equation}
    \theta_i^p =
    \begin{cases}
        \theta_i^p + \pi & \text{if } x_{15}^{\text{kpw}} > x_{16}^{\text{kpw}} \\
        \theta_i^p       & \text{otherwise}
    \end{cases}
\end{equation}
If the right shoulder is positioned to the right of the left shoulder, the person is determined to be facing backward; otherwise, they are facing forward. By utilizing full-body keypoints and PCA-based plane fitting, this method provides a robust and generalizable way to estimate the interaction direction. It is particularly useful for socially aware robots that need to interpret human orientation accurately to engage in human-robot interaction and adaptive navigation. 

\section{Group Interaction Recognition} \label{sec4}
Given the pose and body direction of individuals, our method employs clustering, line extraction, and geometric analysis to identify their interactions.

\subsection{Individual Grouping}
To identify groups of interacting individuals, a clustering algorithm named DBSCAN \cite{kumar2016fast} is used. The algorithm considers a point as part of a cluster if there are at least $N_{\text{min}}$ points within radius $\epsilon$ of the point. Let $P = \{p_1, p_2, \ldots, p_n\}$ be the set of points representing individual positions, $p_i = (x_i^p, y_i^p)$, and $N_\epsilon(p_i) = \{p_j \in P \mid \lVert p_i - p_j \rVert < \epsilon \}$ be the set of the neighborhoods of point $p_i$. Points in $P$ are then grouped based on the following density condition: $|N_\epsilon(p_i)| \geq N_{\text{min}}$.

\subsection{Human Interaction Estimation}
\begin{figure*}[t]
    \centering
    \begin{subfigure}[b]{0.3\linewidth}
        \centering
        \includegraphics[width=\linewidth]{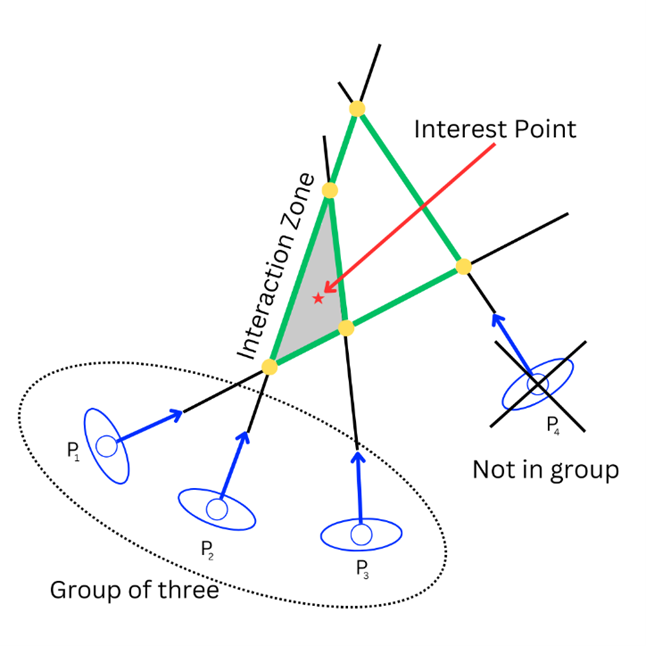}
        \caption{Remove an individual based on $A_{\text{threshold}}$}
        \label{fig:grouping1}
    \end{subfigure}
    \hspace{1cm} % adjust spacing as needed
    \begin{subfigure}[b]{0.3\linewidth}
        \centering
        \includegraphics[width=\linewidth]{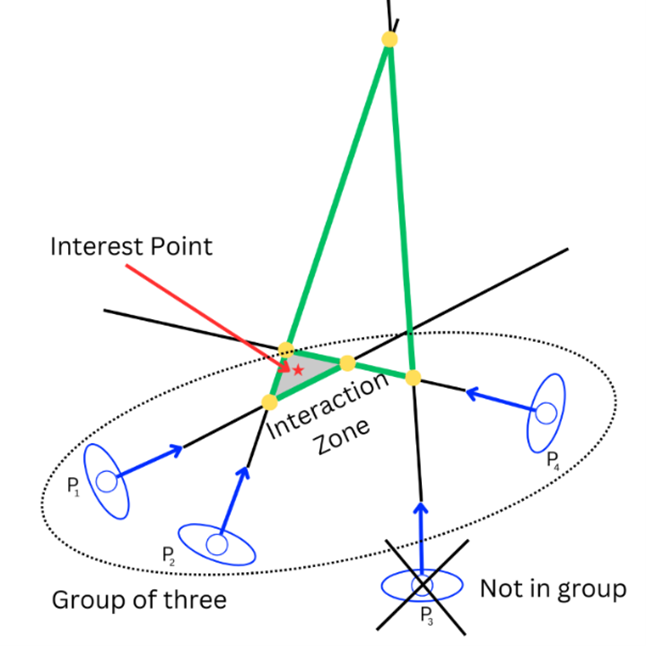}
        \caption{Remove an individual based on $D_{\text{threshold}}$}
        \label{fig:grouping2}
    \end{subfigure}
    \caption{Illustration of individuals grouping and their interactions}
    \label{fig:grouping}
\end{figure*}

To determine whether individuals are interacting, we analyze the relationships between their positions and orientations. As shown in Figure \ref{fig:grouping}, each individual is represented by a line extending from their position, $p_i = (x_i, y_i)$, in the direction of their orientation $\theta_i$. Intersections between lines are then computed to evaluate potential interactions. Denote $L_1$ and $L_2$ as two lines having parametric equations: 

\begin{equation}
    \label{line_equation}
    L_1: p_1 + t_1 d_, L_2: p_2 + t_2 d_2
\end{equation}

where $p_i$ and $d_i$ are the starting point and direction vector of line $i$, and $t_i$ is a scalar parameter. The intersection condition is solved as:

\begin{equation}
    p_1 + t_1 d_1 = p_2 + t_2 d_2
\end{equation}
When intersections exist, a polygon is formed by connecting the intersection points. Area $A$ and center $(x_c^I, y_c^I)$ of this polygon are then calculated from its vertices $(x_i^I, y_i^I)$ using the shoelace formula \cite{lee2017shoelace} as follows:

\begin{equation}
    \label{shoelace_formula}
    A = \frac{1}{2} \sum_{i=1}^{n} \left( x_i^I y_{i+1}^I - y_i^I x_{i+1}^I \right),
\end{equation}

\begin{equation}
    \label{centroid_x_formula}
    x_c^I = \frac{1}{6A} \sum_{i=1}^n (x_i^I + x_{i+1}^I)(x_i^I y_{i+1}^I - y_i^I x_{i+1}^I),
\end{equation}

\begin{equation}
    \label{centroid_y_formula}
    y_c^I = \frac{1}{6A} \sum_{i=1}^n (y_i^I + y_{i+1}^I)(x_i^I y_{i+1}^I - y_i^I x_{i+1}^I).
\end{equation}

To quantify the spatial dispersion of individuals within a group, we define \( D \) as the mean distance from the group centroid to each individual. For \( n \) individuals at positions \((x_i, y_i)\), \( D \) is computed as:

\begin{equation}
    \label{mean_distance}
    D = \frac{1}{n} \sum_{i=1}^n \sqrt{(x_i - x_c^I)^2 + (y_i - y_c^I)^2}.
\end{equation}
This measure provides an estimate of how closely the individuals are positioned relative to the group center. A higher \( D \) indicates a more dispersed group, whereas a lower \( D \) suggests a compact formation. To evaluate if a group has real interactions, metrics \( A \) and \( D \) are used as follows:

\begin{equation}
I = 
\begin{cases}
0, & \text{if } A > A_{\text{threshold}} \text{ or } D > D_{\text{threshold}} \\
1, & \text{otherwise}
\end{cases}
\end{equation}
where \( I = 0 \) indicates weak or non-existent interaction, and \( I = 1 \) indicates significant interaction. The interaction is considered weak if the group's area \( A \) or spread \( D \) exceeds their respective threshold values, suggesting a lack of close engagement among individuals. If a cluster does not meet these conditions, we refine it by identifying individuals not participating in the interaction, as illustrated in Figure 4. The identification is based on the impact on the \( A \) and \( D \) metrics when removing an individual \( p_i \) from the cluster. If removing \( p_i \) results in the new area \( A_i' \) and centroid \( D_i' \) such that:

\begin{equation}
    A_i' \leq A_{\text{threshold}} \quad \text{or} \quad D_i' \leq D_{\text{threshold}}
\end{equation}
Then $p_i$ is permanently removed from the cluster. This process iterates until the cluster meets the social interaction criteria or no significant improvements are obtained. 

\subsection{Algorithm Implementation}
The implementation details of our human interaction recognition method are outlined in the pseudocode below.

\begin{algorithm}[!htbp]
\caption{Group Interaction Recognition}
\label{alg:group-interaction}
\SetAlgoLined
\KwIn{Set of person positions and orientations $\{(x_j^p, y_j^p, \theta_j^p)\}$}
\KwOut{Set of interacting groups $G$}
Let $P = \{(x_j^p, y_j^p)\}$ be positions of all detected persons\;

Apply DBSCAN on $P$ with parameters $\epsilon, N_{min}$ to get clusters $C_k$\;

Initialize empty set $G$\;

\For{each cluster $C_k = \{p_i = (x_i, y_i, \theta_i)\}$}{
    Represent each person as line $L_i$  \tcp*[r]{(Eq.\ref{line_equation})}
    
    Compute all pairwise intersections to form polygon vertices $\{(x_j^I, y_j^I)\}$\;
    
    \If{no polygon formed}{
        Continue\;
    }
    Compute area $A$  \tcp*[r]{(Eq.\ref{shoelace_formula})}
    
    Compute centroid coordinates \tcp*[r]{(Eq.\ref{centroid_x_formula},\ref{centroid_y_formula})}
    
    Compute mean distance D \tcp*[r]{(Eq.\ref{mean_distance})}
    
    \While{$A > A_{threshold}$ or $D > D_{threshold}$}{
        \For{each $p_i$ in $C_k$}{
            Temporarily remove $p_i$, recompute $A_i', D_i'$\; 
            
            \If{$A_i' \leq A_{threshold}$ or $D_i' \leq D_{threshold}$}{
                Permanently remove $p_i$ from $C_k$\;
                
                Update $A \leftarrow A_i'$, $D \leftarrow D_i'$\;
                
                Break\;
            }
        }
        \If{no improvement}{
            Break\;
        }
    }
    \If{$A \leq A_{threshold}$ and $D \leq D_{threshold}$}{
        Add $C_k$ to $G$ as interacting group\;
    }
}
{\raggedright \Return $G$\par}
\end{algorithm}

\section{Results}\label{sec5}
A number of experiments have been conducted to validate the system’s performance in detecting human positions, orientations, and interactions with details as follows.

\subsection{Experimental Setup}
In experiments, a RealSense D435i camera was mounted 0.9 meters above the ground to capture color and depth frames at resolution 1280×720 and 848×480, respectively. Before data collection, the camera’s intrinsic and extrinsic parameters were calibrated to ensure accuracy. The processing unit of the system is a NVIDIA Jetson Orin NX 16GB, an embedded computer typically used for mobile robots. Five scenarios were designed to evaluate a range of interaction dynamics. The first scenario (S1) involves two participants standing 0.8 meters apart and directly facing each other. The second scenario (S2) includes three participants forming a semicircle and facing the camera. The third scenario (S3) arranged three participants in an equilateral triangle. In the fourth scenario (S4), four participants were positioned in a semicircular formation. Finally, the fifth scenario (S5) involved four participants engaged in a circular discussion. Examples of three scenarios are shown in Figure \ref{fig:interaction}. The ground truth data for each scenario was obtained from precise positional and orientation annotations.

\begin{figure*}
    \centering
    \includegraphics[width=0.95\linewidth]{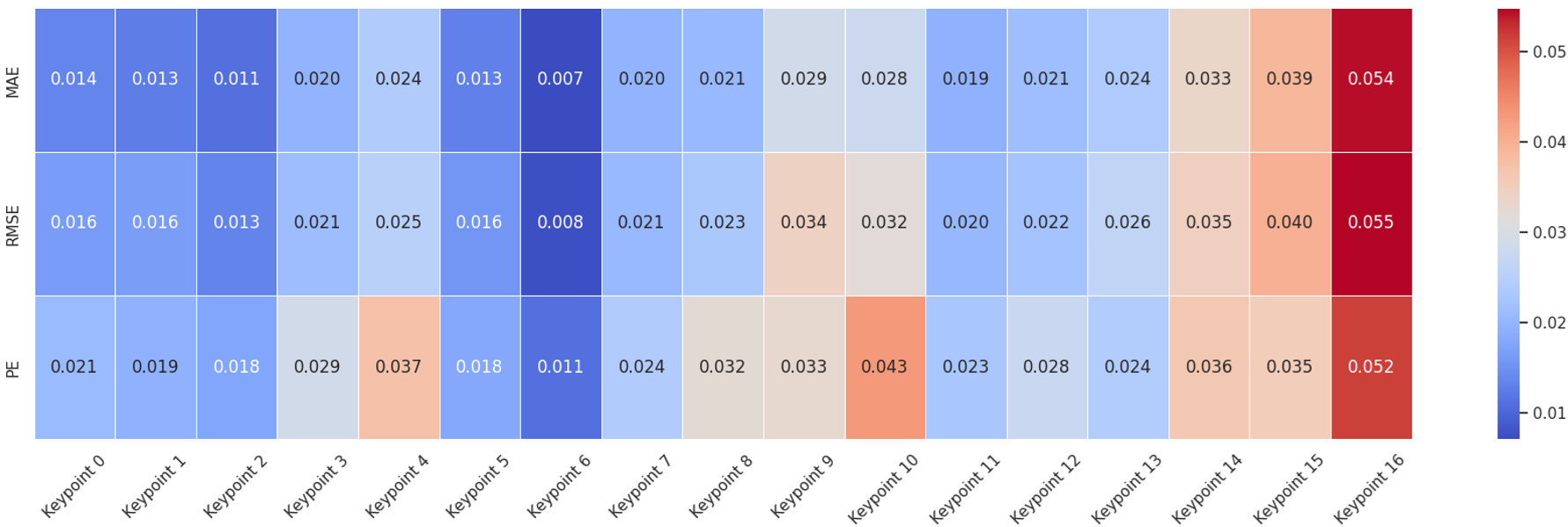}
    \caption{Keypoints estimation errors with three metrics and heatmaps}
    \label{fig:heatmaps}
\end{figure*}

\begin{figure*}[!]
    \centering
    \begin{subfigure}[b]{0.3\textwidth}
    \includegraphics[width=\textwidth]{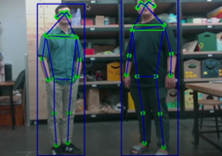}
    \caption{}
    \label{fig:cam}
    \end{subfigure}
    \begin{subfigure}[b]{0.3\textwidth}
    \includegraphics[width=\textwidth]{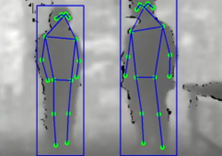}
    \caption{}
    \label{fig:color}
    \end{subfigure}
    \begin{subfigure}[b]{0.3\textwidth}
    \includegraphics[width=\textwidth]{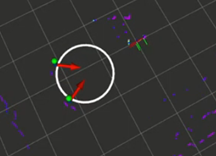}
    \caption{}
    \label{fig:depth}
    \end{subfigure}
    \hfill
        \begin{subfigure}[b]{0.3\textwidth}
    \includegraphics[width=\textwidth]{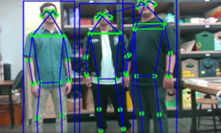}
    \caption{}
    \label{fig:cam}
    \end{subfigure}
    \begin{subfigure}[b]{0.3\textwidth}
    \includegraphics[width=\textwidth]{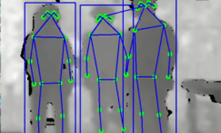}
    \caption{}
    \label{fig:color}
    \end{subfigure}
    \begin{subfigure}[b]{0.3\textwidth}
    \includegraphics[width=\textwidth]{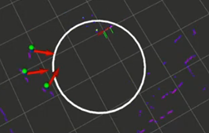}
    \caption{}
    \label{fig:depth}
    \end{subfigure}
    \hfill
    \begin{subfigure}[b]{0.3\textwidth}
    \includegraphics[width=\textwidth]{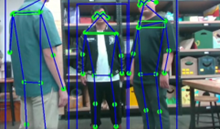}
    \caption{}
    \label{fig:cam}
    \end{subfigure}
    \begin{subfigure}[b]{0.3\textwidth}
    \includegraphics[width=\textwidth]{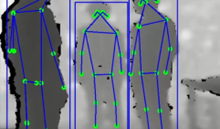}
    \caption{}
    \label{fig:color}
    \end{subfigure}
    \begin{subfigure}[b]{0.3\textwidth}
    \includegraphics[width=\textwidth]{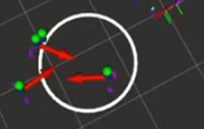}
    \caption{}
    \label{fig:depth}
    \end{subfigure}
    \caption{Human pose and interaction recognition results in different scenarios: (a)(d)(g): detected keypoints in color frames; (b)(e)(h): detected keypoints in depth frames; (c)(f)(i): computed interaction areas on the ground plane}
    \label{fig:interaction}
\end{figure*}

\begin{figure}
    \centering
    \includegraphics[width=0.7\linewidth]{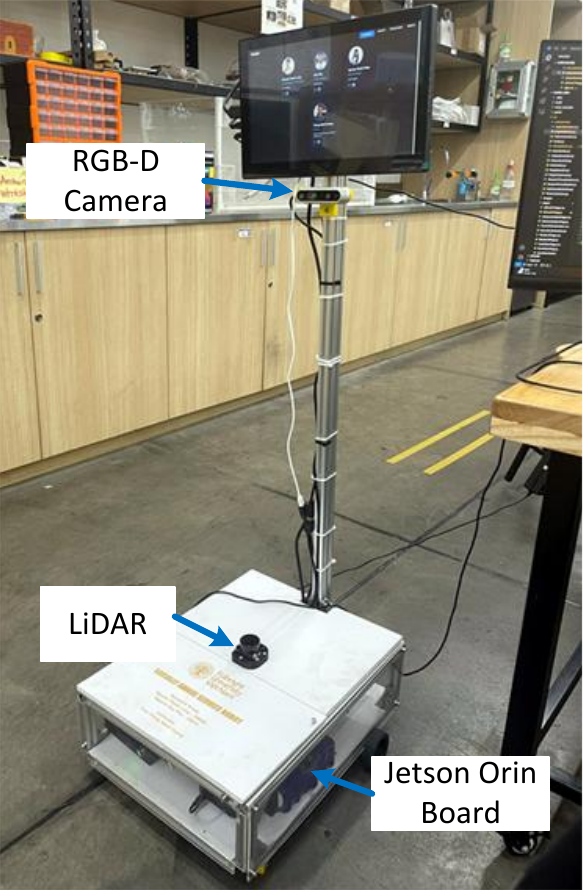}
    \caption{The differential-drive robot equipped with a RealSense D435i camera, a LiDAR sensor, and a Jetson Orin NX computing board for experiments}
    \label{fig:robot}
\end{figure}

\subsection{Human Pose Estimation Result}
In this analysis, the performance of the pose estimator is evaluated by comparing the detected keypoints with the ground truth. Three key metrics including Mean Absolute Error (MAE), Root Mean Squared Error (RMSE), and Percentage Error (PE) are used for evaluation. The results are shown in Figure \ref{fig:heatmaps}. Some keypoints such as Keypoint 6 with MAE = 0.0071, RMSE = 0.0076, and PE = 0.0110 and Keypoint 2 with MAE = 0.0111, RMSE = 0.0133, and PE = 0.0175 have low error, suggesting that the model is particularly strong in predicting these areas. In contrast, keypoints like Keypoint 16 with MAE = 0.0544, RMSE = 0.0548, and PE = 0.0515 and Keypoint 15 with MAE = 0.0390, RMSE = 0.0400, and PE = 0.0353 exhibit higher errors, revealing that the model struggles with predicting more distant points, particularly those on the limbs or extremities. Nonetheless, the overall performance of the pose estimator is sufficiently accurate with an MAE of 0.0229, an RMSE of 0.0249, and a PE of 0.0284.

\begin{table*}[h!]
\centering
\caption{Root mean squared error (RMSE) across five scenarios}
\begin{tabular}{|c|c|c|c|c|c|c|}
\hline
\textbf{Scenario} & \textbf{IZA (m\textsuperscript{2})} & \textbf{IPX (m)} & \textbf{IPY (m)} & \textbf{HPX (m)} & \textbf{HPY (m)} & \textbf{HFD ($^\circ$)} \\
\hline
S1 & 0.002 & 0.023 & 0.037 & 0.023 & 0.063 & 2.489 \\
S2 & 0.021 & 0.075 & 0.123 & 0.087 & 0.102 & 2.987 \\
S3 & 0.043 & 0.032 & 0.087 & 0.092 & 0.024 & 4.423 \\
S4 & 0.038 & 0.089 & 0.186 & 0.123 & 0.129 & 6.897 \\
S5 & 0.067 & 0.124 & 0.087 & 0.154 & 0.086 & 6.942 \\
\hline
\end{tabular}
\label{tab:scenario_metrics}
\end{table*}

\begin{table*}[h!]
\centering
\caption{Time performance of the proposed method}
\begin{tabular}{|l|c|c|c|c|}
\hline
\textbf{Component} & \textbf{2 People (ms)} & \textbf{3 People (ms)} & \textbf{4 People (ms)} & \textbf{Average (ms)} \\
\hline
Pose Estimation & 3.15 & 2.86 & 3.32 & 3.11 \\
Depth Alignment + 3D Localization & 0.55 & 0.52 & 0.55 & 0.54 \\
DBSCAN + Group Detection & 0.25 & 0.14 & 0.15 & 0.18 \\
\hline
\textbf{Total Runtime} & \textbf{3.95} & \textbf{3.52} & \textbf{4.02} & \textbf{3.83} \\
\hline
\end{tabular}
\label{tab:runtime_performance}
\end{table*}

\begin{figure}
    \centering
    \begin{subfigure}[b]{0.7\linewidth}
        \centering
        \includegraphics[width=\linewidth]{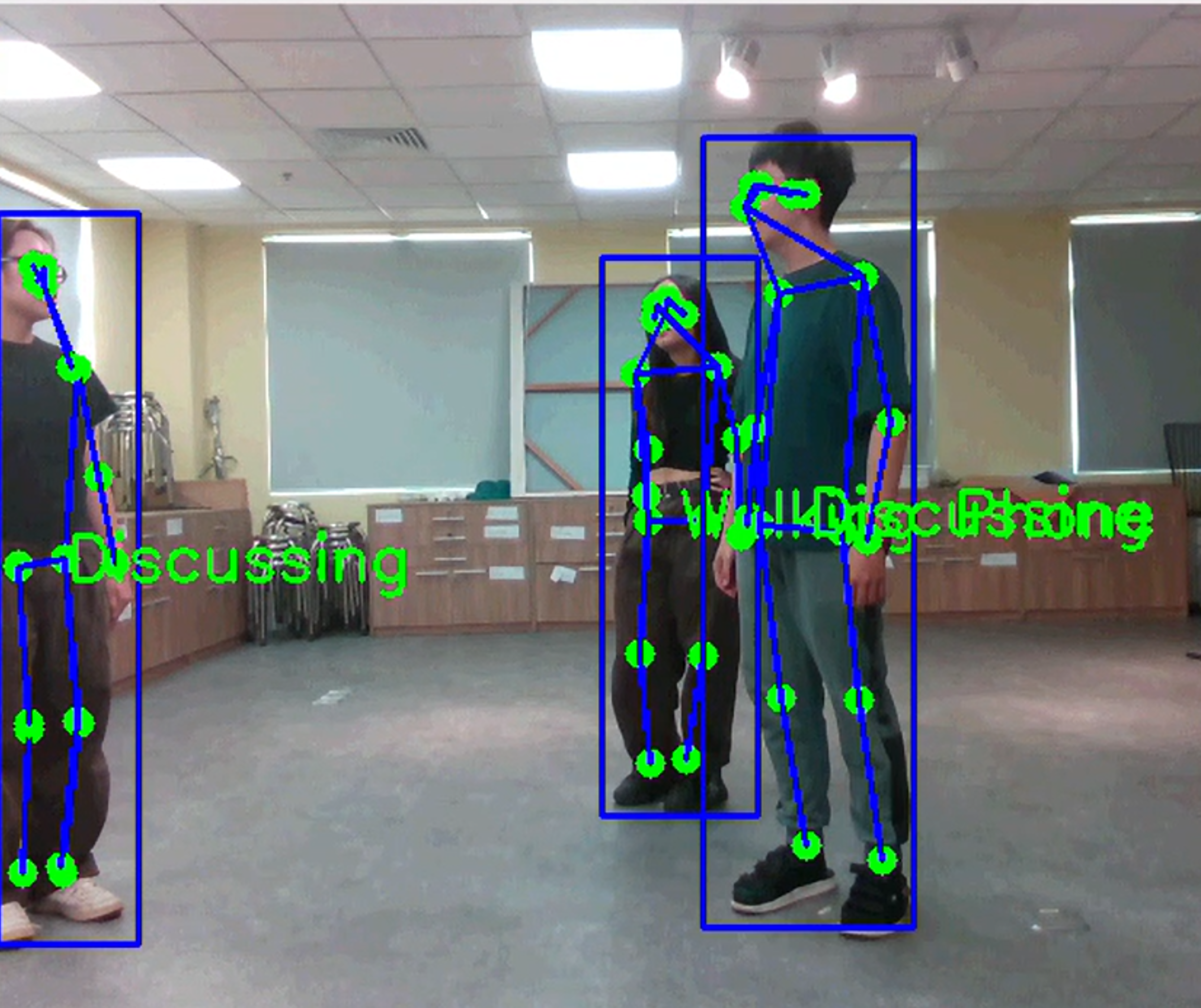}
        \caption{Three people engaged in discussion}
        \label{fig:3people_reg}
    \end{subfigure}

    \vspace{0.5em} % Add some vertical space between subfigures

    \begin{subfigure}[b]{0.7\linewidth}
        \centering
        \includegraphics[width=\linewidth]{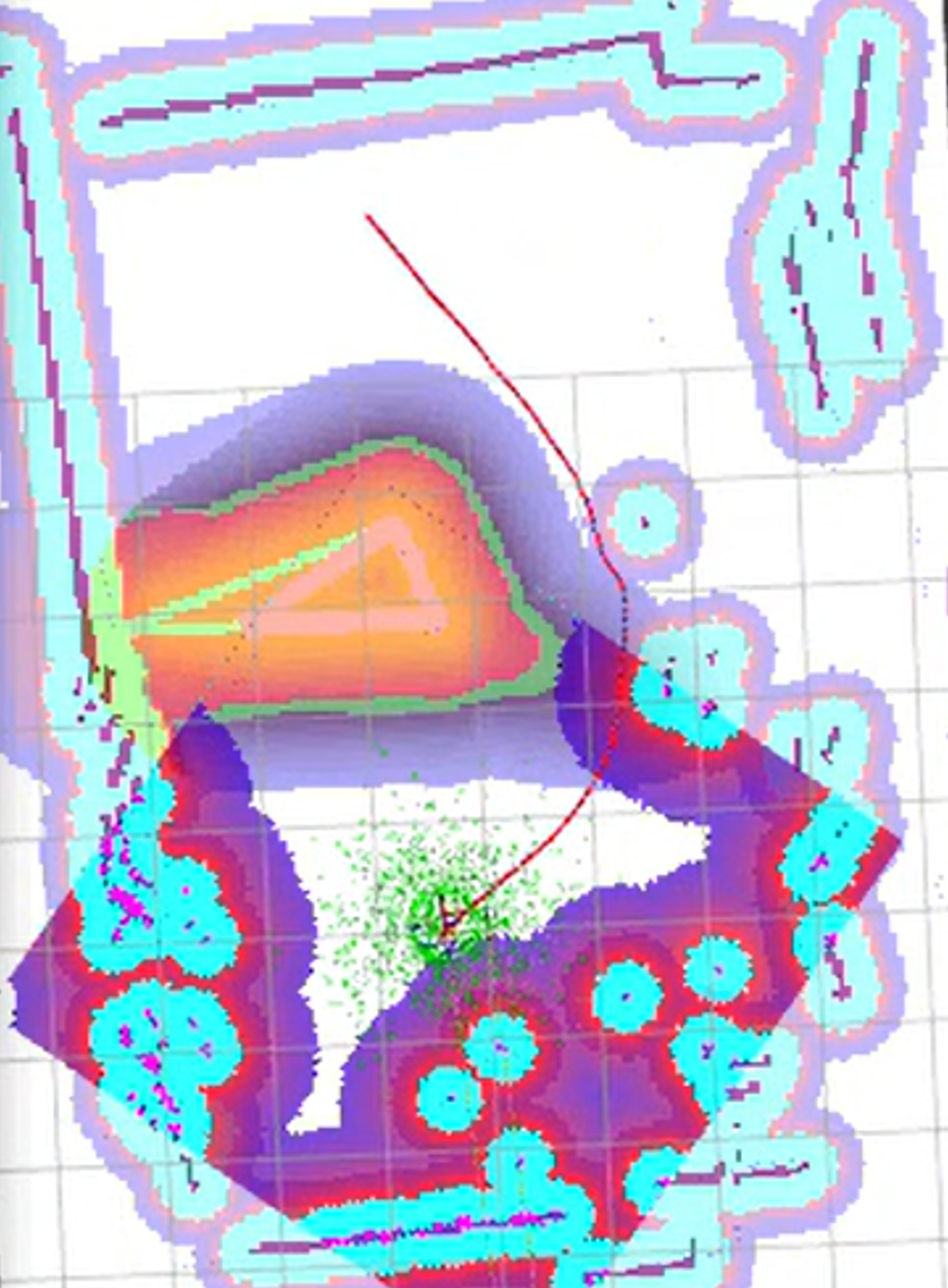}
        \caption{Navigation costmap showing the group interaction space (orange area)}
        \label{fig:costmap}
    \end{subfigure}

    \caption{Interaction recognition results and associated costmap for a group of three individuals engaged in interaction}
    \label{fig:3people}
\end{figure}

\begin{figure*}[htp]
    \begin{subfigure}{\textwidth}
        \includegraphics[width=\textwidth]{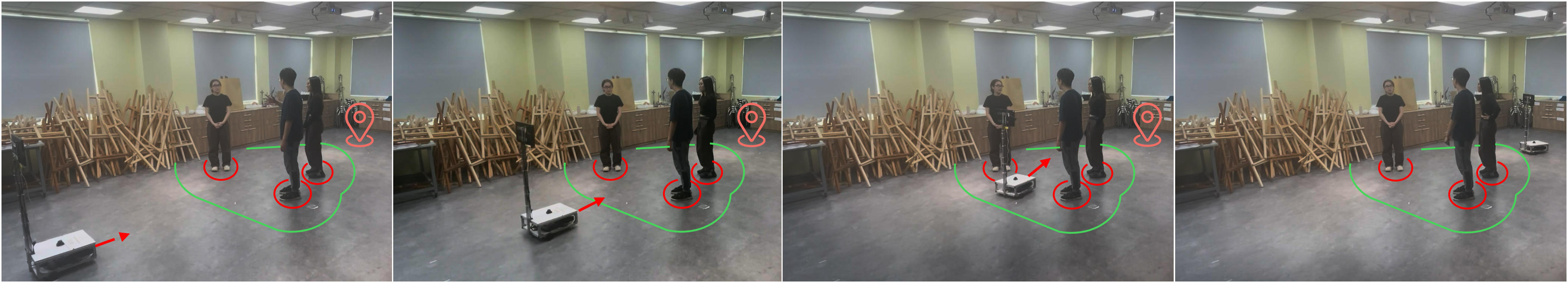}
        \caption{Robot motion in the absence of human group interaction recognition}
        \label{fig:nav1}
    \end{subfigure}
    \begin{subfigure}{\textwidth}
        \includegraphics[width=\textwidth]{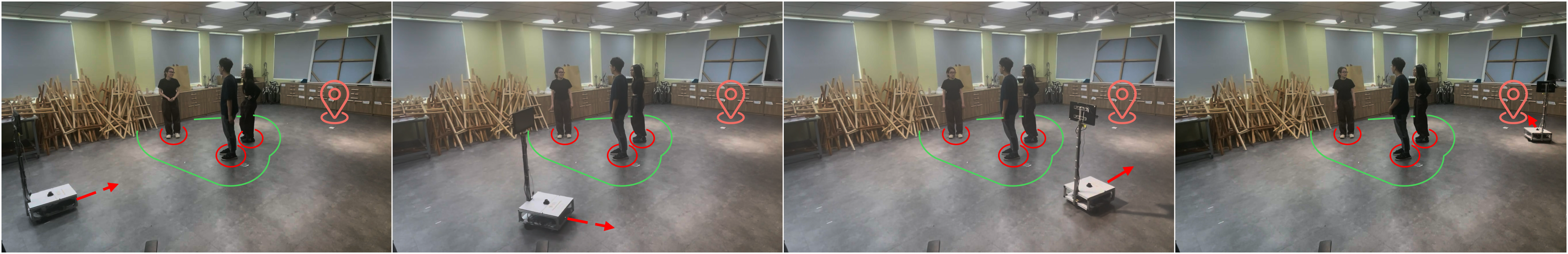}
        \caption{Robot motion in the presence of human group interaction recognition}
        \label{fig:nav2}
    \end{subfigure}
    \caption{Robot navigation results}
    \label{fig:nav}
\end{figure*}

\subsection{Group Detection Results}
Figure \ref{fig:interaction} shows the group detection results. The individuals are successfully detected and grouped in different scenarios. Their interaction areas are also computed and projected on the ground plane, as shown in Figure \ref{fig:interaction}c, f, i. To further evaluate the results, we computed the root mean squared error (RMSE) across five scenarios for key interaction parameters, including interaction zone area (IZA), interaction point x (IPX), interaction point y (IPY), human position x (HPX), human position y (HPY), and human facing direction (HFD). 

Table \ref{tab:scenario_metrics} shows the detection results where the system maintains relatively high accuracy in estimating human positions, orientations, and interaction zones across all scenarios. The RMSE for HPX, HPY, IPX, IPY remains within 20 cm range, which is sufficient for most robot navigation tasks. As the number of individuals increases, the RMSE for HFD and IZA slightly increases because of the decrease in the number of features available to detect each individual and the appearance of partial occlusions. The results reflect the challenge of accurately identifying a central interaction point when multiple individuals are engaged.

\subsection{Real-time Performance Evaluation}
The real-time performance is evaluated to assess the suitability of the proposed method for practical robot navigation. Table \ref{tab:runtime_performance} shows the runtime for the scenarios with 2, 3, and 4 participants. This runtime accounts for the entire processing pipeline, including pose estimation, depth alignment, body direction estimation, and interaction detection. One key observation from the runtime analysis is that increasing the number of participants does not significantly affect the overall processing time. The system maintains a near-constant processing speed across all tested scenarios. This result highlights the efficiency of the framework in handling multiple people without a proportional increase in computational load.

Pose estimation remains the most computationally intensive step, averaging 3.11 ms per frame. Depth alignment and 3D localization requires 0.54 ms on average. Interaction detection is the least computationally expensive step, averaging only 0.18 ms. Overall, the framework demonstrates robust real-time performance when processing each frame in under 4.1 ms across all scenarios. The near-constant runtime across different number of participants confirms the scalability of the system, making it suitable for real-world socially aware robotic applications where multiple individuals interact in shared spaces.

\subsection{Robot Navigation Validation}
To evaluate the validity of the proposed human interaction recognition (HAR) method for socially-aware robot navigation, experiments were conducted using a differential-drive mobile robot, as shown in Figure ~\ref{fig:robot}. The robot is equipped with a LiDAR sensor, an RGB-D camera, and a single-board computer for autonomous operation. The LiDAR has a 360$^\circ$ scanning angle and a 10-meter range, and is responsible for generating the base costmap. The camera, a RealSense D435i, is mounted at a height of 1 m and has a field of view (FoV) of 87$^\circ$, used for detecting and analyzing human activities. The single-board computer, a Jetson Orin NX 16GB, runs the Robot Operating System 2 (ROS 2). It manages the robot's software stack and real-time processing.

The proposed recognition algorithm is implemented as a ROS 2 package that outputs the recognized human interaction region as a social costmap layer. This layer is then integrated with the LiDAR-generated costmap to produce the final navigation costmap.

Figure~\ref{fig:nav} illustrates an experimental scenario in which the robot navigates through a group of three people engaged in a conversation. Without the recognition module, the robot perceives the individuals as isolated obstacles and attempts to pass through the space between them, as shown in Figure \ref{fig:nav1}. When the recognition module is enabled, the robot recognizes their group interaction, even when one person is partially occluded, as shown in Figure \ref{fig:3people_reg}. The resulting social interaction space is incorporated into an adaptive costmap, represented by the orange area in Figure \ref{fig:costmap}. This allows the robot to plan a path (the red line) that avoids the entire group. As a result, the robot navigates to the goal without disrupting the group interaction, as illustrated in Figure \ref{fig:nav2}. It shows that the recognition module enables the robot to better understand the contextual relationships between humans and their environment and adapt its behavior to respect social norms. This leads to more socially appropriate navigation that maintains not only physical safety but also psychological comfort for humans.

\section{Conclusion}\label{sec6}
In this work, we have presented a novel method for recognizing human interactions using a single RGB-D camera. The method is capable of recognizing human activities and interactions under various conditions, including partial occlusion. It achieves high-speed performance of approximately 4 ms per frame on resource-constrained hardware, such as single-board computers used in robotic systems. The recognition is accurate and scalable to human groups of varying sizes. The implementation is carried out as a ROS 2 package, making it easy to integrate into existing navigation platforms. 

In future work, we aim to improve the system’s accuracy under challenging conditions, such as when individuals are oriented $\pm90^\circ$ relative to the camera, to make it even more reliable and versatile for socially-aware robot navigation.

\balance

\bibliographystyle{ieeetr}  
\bibliography{ref}

\end{document}